%% file: main.tex
\documentclass[conference]{IEEEtran}
\IEEEoverridecommandlockouts
\usepackage{booktabs}
\usepackage{cite}
\usepackage{amsmath,amssymb,amsfonts}
\usepackage{algorithmic}
\usepackage{graphicx}
\usepackage{textcomp}
\usepackage{xcolor}
\def\BibTeX{{\rm B\kern-.05em{\sc i\kern-.025em b}\kern-.08em
    T\kern-.1667em\lower.7ex\hbox{E}\kern-.125emX}}
\begin{document}
\title{Holistix: A Dataset for Holistic Wellness Dimensions Analysis in Mental Health Narratives\\
}

\author{\IEEEauthorblockN{Heba Shakeel}
\IEEEauthorblockA{\textit{Department of Computer Engineering} \\
\textit{Jamia Millia Islamia}, New Delhi, India  \\
heba1907468@st.jmi.ac.in}
\and
\IEEEauthorblockN{Tanvir Ahmad}
\IEEEauthorblockA{\textit{Department of Computer Engineering} \\
\textit{Jamia Millia Islamia}, New Delhi, India \\
tahmad2@jmi.ac.in}
\and
\IEEEauthorblockN{ Chandni Saxena}
\IEEEauthorblockA{\textit{The Chinese University of Hong Kong} \\
Hong Kong SAR, China \\
csaxena@cse.cuhk.edu.hk}

}

\maketitle

\begin{abstract}
We introduce a dataset for classifying wellness dimensions in social media user posts, covering six key aspects: physical, emotional, social, intellectual, spiritual, and vocational. The dataset is designed to capture these dimensions in user-generated content, with a comprehensive annotation framework developed under the guidance of domain experts. This framework also includes labeling text spans from these posts to provide explanations that highlight the corresponding wellness aspects. We evaluate both traditional machine learning models and advanced transformer-based models for this multi-class classification task, with performance assessed using precision, recall, and F1-score, averaged over 10-fold cross-validation. Post-hoc explanations are applied to ensure the transparency and interpretability of model decisions. The proposed dataset contributes to region-specific wellness assessments in social media and paves the way for personalized well-being evaluations and early intervention strategies in mental health.  We adhere to ethical considerations for constructing and releasing our experiments and dataset publicly on Github~\footnote{https://github.com/HebaShakeel/holistix}.
\end{abstract}

\begin{IEEEkeywords}
Mental Health, Social Media, User Posts, Dataset, Wellness Dimensions.
\end{IEEEkeywords}
\input{Introduction}
\input{Dataset}
\input{ExpAndEvaluation}

\input{LimConclusion}
\input{EthicalStmt}
\section*{Acknowledgment}
We would like to thank Dr. Muskan Garg (Mayo Clinic, Rochester USA)  for her invaluable input and guidance in shaping the foundation of this work. Special thanks to our collaborator, Zahra Tausif, a Cognitive Neuropsychologist (Riyadh, Saudi Arabia), for providing the annotation guidelines and wellness dimension indicators for text segments. We extend our gratitude to the student annotators whose contributions were essential to this research. We also acknowledge Tanzeel Mujtaba for his valuable support at various stages of this project.
\vspace{12pt}
\input{main.bbl}

\end{document}

%% file: Introduction.tex
\section{Introduction}
Mental health disorders have become a critical global health problem, with more than a billion people worldwide affected by mental, neurological, and substance use disorders~\cite{whoWorldMental}. Depression alone impacts approximately 280 million people and stands as a leading cause of disability globally~\cite{whoDepressiveDisorder}. Despite the prevalence of these conditions, access to timely and effective mental health care remains a significant challenge, often due to stigma, resource limitations, and a shortage of mental health professionals~\cite{henderson2013mental}. 

In parallel, the advent of social media has transformed communication, with platforms like Facebook, Twitter, and Instagram becoming integral to daily life. While these platforms offer avenues for connection and information sharing, excessive use has been linked to negative mental health outcomes, including increased anxiety and depression~\cite{mitStudySocial}. Conversely, social media also provides a unique opportunity for mental health assessment and intervention~\cite{mangalik2024robust}. Individuals often share their thoughts and emotions online, creating a rich data source that, with appropriate analysis, can offer an understanding of population-level mental health trends. 
\begin{figure}[tbp]
\centering
\includegraphics[width=\columnwidth]{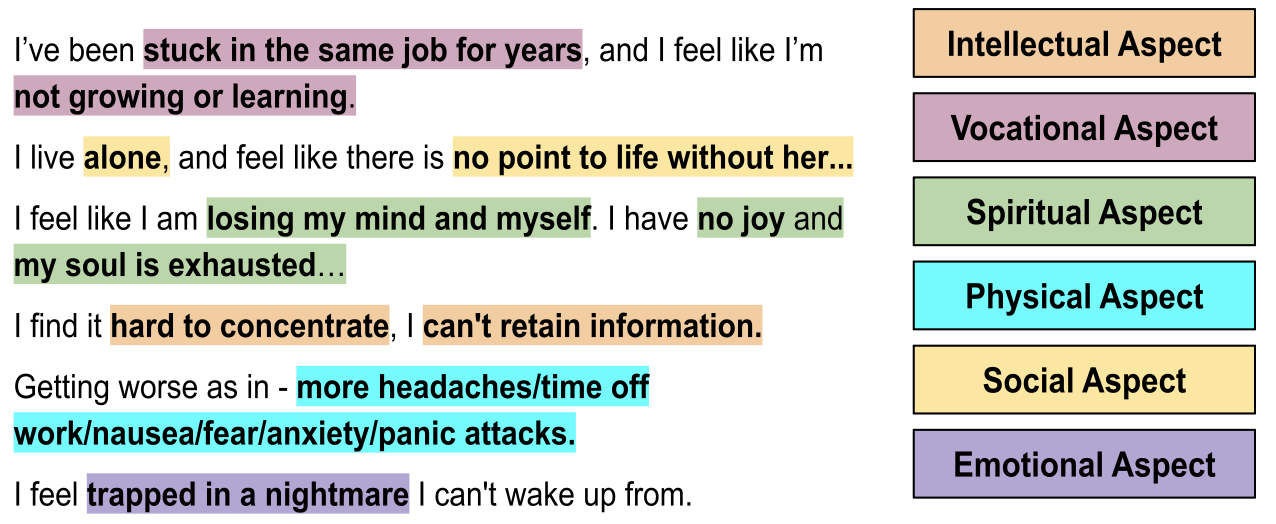}
\caption{Overview of the problem formulation showing the identification of wellness dimensions in user posts.}
\label{fig}
\end{figure}
Studies have demonstrated the potential of AI and NLP techniques to analyze social media content, enabling monitoring of mental health indicators across diverse populations~\cite{le2021machine, garg2023nlp}. By leveraging the vast amount of user-generated content, researchers and practitioners can develop innovative, non-intrusive methods for early detection, continuous assessments, and personalized interventions~\cite{mansoor2024early}. 

Despite the potential of AI and NLP techniques to analyze social media content for mental health monitoring, several challenges persist. One significant issue is the scarcity of high-quality, annotated datasets that capture the multifaceted nature of mental health expressions across diverse populations~\cite{garg2023mental}. Existing datasets often lack the granularity required to identify specific wellness dimensions, such as emotional, social, and psychological well-being, which are crucial for comprehensive mental health assessments.

To address this gap, we introduce a fine-grained approach to mental health assessment that frames the task as wellness concept classification accompanied by explanatory text span annotations. This involves creating a novel, population-specific ``Holistix" dataset focusing on the Australian population, with data collected from users reporting on a mental health forum\footnote{https://www.beyondblue.org.au}. Our dataset comprises 1420 instances totaling 37082 words, annotated by experts following a structured annotation scheme grounded in Halbert L. Dunn’s theory of wellness dimensions. Fig. 1 illustrates the potential for tailoring mental health assessments by analyzing human writings through the lens of wellness dimensions, thereby moving beyond basic evaluations. Notably, while previous studies, such as~\cite{garg2024wellxplain}, utilized Reddit data and focused on four of Dunn's wellness dimensions, our research advances this by encompassing all six dimensions: physical, emotional, intellectual, spiritual, social, and vocational aspects~\cite{stara2013wellness}. We refer to the six wellness dimensions as `aspects' to maintain consistency in terminology with recent works~\cite{garg2024wellxplain,mohammadi2024welldunn}. This comprehensive approach provides a more holistic understanding of mental health as expressed in social media contexts. We fine-tune baseline transformer-based models and use traditional machine learning techniques for multi-class wellness dimension classification. The proposed dataset contributes to region-specific wellness assessments in social media and paves the way for personalized well-being and early mental health intervention.

%% file: Dataset.tex
\section{Holistix Dataset}
The utilization of social media data has become pivotal in understanding mental health trends and behaviors. Platforms like Facebook~\cite{reece2017instagram}, Twitter~\cite{xue2020public}, Reddit~\cite{haque2021deep}, and region-specific platforms like Weibo~\cite{cheng2017assessing} offer rich, user-generated content that researchers analyze to gain insights into public sentiment and mental health issues. Although some data resources are publicly available, many datasets remain inaccessible due to concerns over user privacy and adherence to ethical standards.

Reddit, with its diverse and anonymous user base, provides valuable data for mental health research~\cite{cohan2018smhd,turcan2019dreaddit,haque2021deep,garg2022cams,garg2024wellxplain}, while platforms like Weibo~\cite{cheng2017assessing} (a Chinese microblog),  offer localized insights into region-specific mental health discussions. Comparative analyses of global data from Reddit and country-specific data from Reddit~\cite{rai2024cross} reveal significant cross-cultural differences in mental health expressions, underscoring the need for culturally sensitive research and interventions. Despite existing studies, there remains a critical need for more country-specific datasets to capture cultural, social, and linguistic nuances influencing mental health. To address this gap, we have curated the Holistix dataset from Australia's Beyond Blue~\cite{beyondblue247Support} mental health forums, reflecting the unique challenges faced by the Australian population. This data will help develop targeted, culturally appropriate interventions and enrich mental health research with localized perspectives.
\subsection{Dataset Creation} 
Beyond Blue~\cite{beyondblue247Support}, one of Australia’s most reputable mental health organizations, provides a secure and anonymous platform for individuals to access information, seek advice, and share experiences. The anonymity fosters open discussions, making it a more viable resource for mental health research compared to conventional social media platforms. We extracted 2,000 raw posts from Beyond Blue’s online forums using the Python library- BeautifulSoup\footnote{https://pypi.org/project/beautifulsoup4/}. The data was sourced from key discussion categories, including \textit{Anxiety}, \textit{Depression}, \textit{PTSD and Trauma}, \textit{Suicidal Thoughts and Self-Harm}, \textit{Relationship and Family Issues}, \textit{Supporting Friends and Family}, and \textit{Grief and Loss}. To ensure user privacy, only the textual content and its corresponding category were retained.

To enhance data quality, we performed rigorous preprocessing by removing irrelevant, empty, and duplicate posts. Posts that were excessively long or off-topic were also filtered out, refining the dataset to 1,420 posts specifically focused on mental distress. 
\subsection{Annotation Framework}
For accurate classification, domain experts that include a social NLP researcher and a senior clinical psychologist who developed structured annotation guidelines. They collaboratively defined clear class indicators,  annotation schemes, and perplexity criteria to categorize posts based on underlying wellness dimensions, ensuring consistency and reliability in the labeling process. 
\subsubsection{Wellness Dimensions}
Building upon the foundational work of Halbert L. Dunn and Dr. Bill Hettler, our research focuses on the intersection of these wellness dimensions and mental health within the Australian context. Halbert L. Dunn introduced the concept of high-level wellness~\cite{stara2013wellness} in the 1950s, emphasizing a holistic approach to health beyond the mere absence of disease. Building upon Dunn's foundational work, Bill Hettler~\footnote{https://www.hettler.com/history/hettler.htm}  developed the Six Dimensions of Wellness model. This model comprises the following dimensions: 
\begin{itemize}
    \item \textbf{Intellectual Aspect.} Encourages engaging in creative and stimulating activities to expand knowledge and skills.
    \item \textbf{Vocational Aspect.} Involves personal satisfaction and enrichment derived from one's work, contributing meaningfully to society.
    \item \textbf{Spiritual Aspect.} Entails seeking purpose and meaning in human existence, leading to a harmonious life.
    \item \textbf{Physical Aspect.} Recognizes the need for regular physical activity, healthy dietary choices, and preventive health measures.
    \item \textbf{Social Aspect.} Emphasizes developing a sense of connection and belonging through positive interpersonal relationships.
    \item \textbf{Emotional Aspect.} Involves awareness and acceptance of one's feelings, coping effectively with stress, and maintaining satisfying relationships.
\end{itemize}

\subsubsection{Annotation Process}
\begin{figure*}[htbp]
\centering
\includegraphics[width=0.8\textwidth]{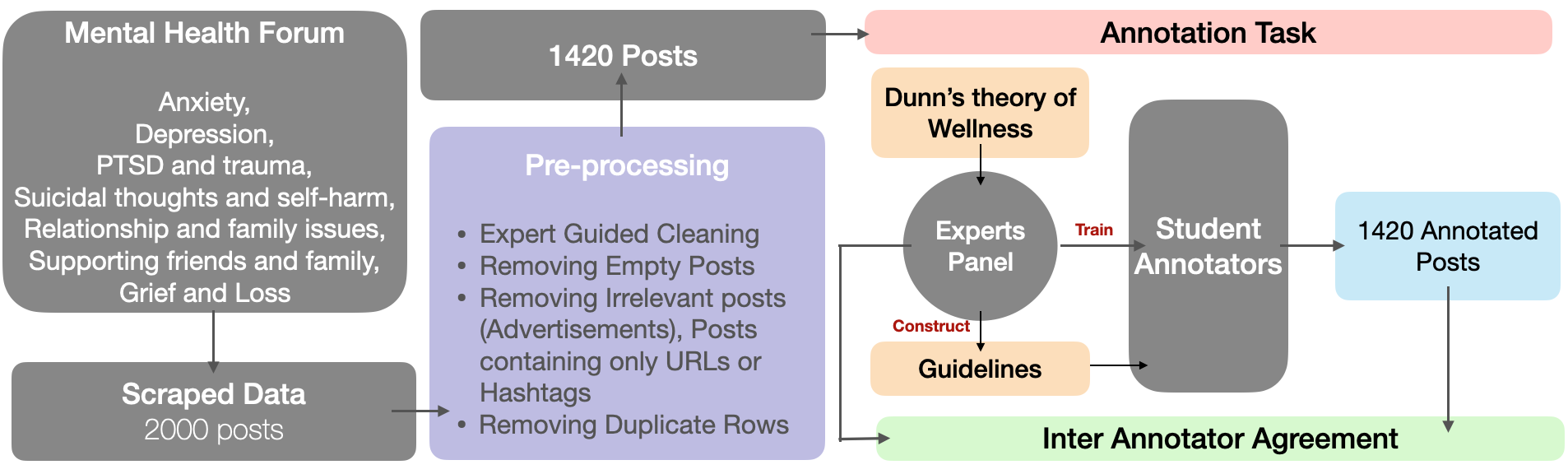}
\caption{Framework of data annotation process}
\label{fig:framework}
\end{figure*}

As shown in Fig.~\ref{fig:framework}, the framework highlights the annotation process, detailing the steps involved in generating and labeling the data. This includes the incorporation of guidelines referring to Dunn's model\footnote{https://cdn.ymaws.com/members.nationalwellness.org/resource/resmgr/\\cwp/sixdimensions\_overview\_asses.pdf}, the integration of expert feedback, and the use of the annotation framework for wellness dimensions. Table~\ref{tab:indicator} provides the indicators used to identify the wellness dimension class within text spans, assisting annotators in accurately classifying the relevant dimensions based on predefined criteria.
\begin{table*}[htbp]
\caption{Class indicators in social media posts for the annotation task, where class indicates the Wellness dimension}
\label{Tab:class_indicators}
\begin{center}
\begin{tabular}{|p{0.5cm}|p{10cm}|p{5cm}|}
\hline
\textbf{Class} & \textbf{ Indicators in text} & \textbf{Example} \\
\hline
\textbf{PA} &Mentions of fatigue, sleep issues, body image concerns, diet struggles, illness, or medication. Phrases related to body shaming, physical deterioration, weight concerns, or health anxiety.
&\textit{“I feel exhausted all the time and can’t even sleep properly.” Or “I hate my body and feel disgusting when I look in the mirror.”}\\
\hline
\textbf{IA}
 &Discussions about academic stress, feelings of intellectual inadequacy, frustration with learning.
&\textit{“I feel like I’ll never be smart enough to pass my exams.”}\\
\hline
\textbf{VA}
& Workplace dissatisfaction, career struggles, financial burdens related to work or dissatisfaction with career progression. &\textit{“My 9-5 job drains me, and I don’t see the point in trying anymore.”}\\
\hline
\textbf{SA}
 &Mentions of loneliness, strained relationships, loss of social support, feeling excluded or isolated. Discussions about family, friends, breakups, bullying, or lack of belonging.
&\textit{“I have no real friends, and I feel invisible at school.”} Or \textit{“Ever since my breakup, I feel like I’ve lost my entire social circle.”}\\
\hline
\textbf{SpiA}
 &Expressions of hopelessness, self-doubt, existential crises, or struggling with purpose in life.
&\textit{“I don’t know what my purpose is anymore, and everything feels meaningless.”}\\
\hline
\textbf{EA}
 &Emotional instability, feelings of emotional exhaustion, inability to cope, or extreme sadness.
&\textit{“I hate myself and don’t think I belong in this world.”}\\
\hline
\end{tabular}
\label{tab:indicator}
\end{center}
\end{table*}
\subsection{Statistics of the Dataset}
We analyzed the distribution of samples across different aspects: Intellectual Aspect (IA) – 10.91\%, Vocational Aspect (VA) – 10.56\%, Spiritual Aspect (SpiA) – 13.38\%, Physical Aspect (PA) – 20.84\%, Social Aspect (SA) – 28.59\%, and Emotional Aspect (EA) – 15.70\%. Table~\ref{Tab:Stats} presents the statistical overview of the dataset.
\begin{table}[htbp]
\caption{Statistics of Dataset}
\label{Tab:Stats}
\begin{center}
\begin{tabular}{|c|c||c|c|}
\hline
\textbf{Measure} & \textbf{Count} &\textbf{Wellness Dimension} & \textbf{Count} \\
\hline
Total posts &1420 &\textbf{IA} &155\\
Total words count &37082 &\textbf{VA} &150\\
Max. word count per post &115 &\textbf{SpiA} &190\\
Total sentence count &2271 &\textbf{PA} &296\\
Max. sentences per post &9 &\textbf{SA} &406\\
 & &\textbf{EA} &223\\
\hline
\end{tabular}
\end{center}
\end{table}
The high percentage of samples in the Social Aspect (SA) suggests that many individuals in Australian society struggle with issues related to relationships, loneliness, and emotional connection. These challenges can be complex and difficult to navigate, requiring significant attention and resources to address. The large volume of posts in this category also indicates that people are actively seeking emotional support and validation through online platforms, highlighting the growing role of social media in mental health discussions.
On the other hand, the relatively lower number of samples in the IA and VA may indicate that these areas are well-regulated and efficiently managed. This could be due to strong governance, established policies, or the expertise of professionals within these domains, reducing the frequency of related concerns being expressed online. The extracted text spans represent key portions of user posts that justify the classification into specific wellness aspects. These spans capture the contextual cues and underlying themes that indicate a user's mental health concerns. By analyzing these explanatory segments, we identify the most frequently occurring words associated with each wellness aspect, providing insights into common linguistic patterns and themes. The results are systematically presented in  Table~\ref{Tab:freq_words} for further interpretation.

\begin{table*}[htbp]
\caption{Frequent Words in Explainatory Text Spans Across Wellness Dimensions}
\label{Tab:freq_words}
\begin{center}
\begin{tabular}{|c|c|}
\hline
\textbf{Wellness Dimension} & \textbf{Most Frequent Words (Avg. Count)} \\
\hline
Intellectual Aspect &future(10), feel(9), hard(9), thoughts(7), lack(7), think(6), struggling(5) \\
\hline
Vocational Aspect
&job(45), work(43), money(8), career(7), financial(7), struggling(6), unemployed(6)\\
\hline
Spiritual Aspect
&feel(40), life(31), thoughts(9), suicide(8), struggling(7), feeling(6)\\
\hline
Social Aspect
&me(48), people(35), feel(43), talk(21), alone(18), friends(17), relationship(17)\\
\hline
Physical Aspect &anxiety(42), sleep(30), depression(28), disorder(17), diagnosed(14), bad(11)\\
\hline
Emotional Aspect
&feel(41), anxiety(23), feeling(18), me(9), sad(8), crying(7), hard(7)\\
\hline
\end{tabular}
\label{tab1}
\end{center}
\end{table*}
\subsection{Data Annotation and Perplexity}
The annotation and perplexity guidelines aim to provide structured frameworks to help annotators consistently interpret complex text. 
The rest of this section covers the rules for the annotation process, including  text span, categorization, and perplexity handling.
\subsubsection{Data Annotation Guidelines}
The following guidelines offer a set of rules and recommendations to assist annotators in navigating complex or ambiguous text, ensuring that each annotation accurately reflects the intended wellness dimension:
\begin{enumerate}
 \item Identify relevant text spans in the posts (words or phrases that describe thoughts, actions or feelings linked to a wellness dimension)
 \item Handling overlaps (initially label all the relevant dimensions if a text span fits multiple dimensions, later based on perplexity guidelines, the key dimension will be determined and assigned)
 \item Annotations should be specific, i.e., the exact words or phrases should be highlighted that indicate a wellness dimension. 
 \item If the post is very lengthy, focus on text that shows how the dimension impacts mental well-being.
 \item Only annotate what is explicitly stated or strongly implies, i.e. avoid assumptions.
 \item Each annotated text entry should include: 
 \begin{itemize}
     \item The text: user’s social media post
    \item Text span: the key phrases in the text
    \item Wellness dimension: one of the six labels (dimensions)
 \end{itemize}
  \item Determine the quality of the annotations:
   \begin{itemize}
    \item This can be done by having a second annotator review 20\% of the entries to ensure consistency.
    \item Discussing the ambiguous cases; will help to refine the guidelines.
    \end{itemize}
\end{enumerate}
\subsubsection{Perplexity Guidelines}
The perplexity guidelines are designed to support the annotation process by offering rules and suggestions for navigating these challenges, ensuring consistency and accuracy in tagging complex content. The key principles include:
\begin{enumerate}
    \item \textbf{Prioritize Dominant Dimensions.} If a text spans \textbf{multiple wellness dimensions}, label \textbf{all relevant ones but highlight the most dominant} (based on context or emphasis).
    \begin{itemize}
        \item Example:
        \begin{itemize}
            \item Text: \textit{“My volunteer work (Vocational) helps me connect with others (Social), but  I’m exhausted (Physical).”}
            \item Labels: \textbf{Vocational} (dominant), \textbf{Social, Physical}.
        \end{itemize}
    \end{itemize}
    \item \textbf{Resolve Ambiguity with Context Clues.} If the meaning is unclear, use surrounding sentences to infer the dimension.
    \begin{itemize}
        \item Example:
        \begin{itemize}
            \item Text: \textit{“I feel overwhelmed.”}
            \item Context: Previous sentence mentions \textit{“my boss gave me three deadlines.”} $\rightarrow$ Label \textbf{Vocational}.
        \end{itemize}
    \end{itemize}
    \item \textbf{Break Down Compound Sentences.} Split sentences with \textbf{multiple independent clauses} into separate annotations.
    \begin{itemize}
        \item Example:
        \begin{itemize}
            \item Text: \textit{“I journal to manage stress (Emotional), but my poor diet (Physical) isn’t helping.”}
            \item Split into two entries.
        \end{itemize}
    \end{itemize}
    \item \textbf{Avoid Overinterpreting Metaphors/Sarcasm.} Label metaphors or sarcasm \textbf{literally} unless the tone is obvious.
    \begin{itemize}
        \item Example:
        \begin{itemize}
            \item Text: \textit{“Oh yeah, my ‘healthy’ routine of 2 hours of sleep is working great!”}
            \item Label: \textbf{Physical} (negatively impacted)
        \end{itemize}
    \end{itemize}
    \item \textbf{Label Implicit Meanings Sparingly.} Only label \textbf{implied wellness aspects} if strongly supported by context. Avoid guessing.
    \begin{itemize}
        \item Example:
        \begin{itemize}
            \item Text: \textit{``I haven’t left my room in days.”}
            \item Implicit labels: \textbf{Social} (isolation), \textbf{Physical} (inactivity)
        \end{itemize}
    \end{itemize}
    \item \textbf{Validate Annotations with Team Consensus.} Discuss 10\% of ambiguous cases as a team to align interpretations. Update guidelines based on recurring dilemmas.
\end{enumerate}
\subsection{Annotation Task}
Based on the guidelines curated by our team, we train two student annotators for the task of 6-class classification and text-span identification. To ensure the reliability of the annotations, we asked the student annotators to annotate the data instances independently. These annotations underwent a statistical evaluation using Fleiss’ Kappa~\cite{mchugh2012interrater} interobserver agreement, yielding a value of $\kappa= 75.92\%$. 

%% file: ExpAndEvaluation.tex
\section{Experiments and Evaluation}
We establish baselines using traditional machine learning and transformer models for multi-class classification. We used a fixed set of $990$ training samples, $212$ validation samples, and $213$ test samples, with performance evaluated using 10-fold cross-validation. After evaluating performance, we apply LIME post-hoc explanations to the best-performing model to enhance interpretability.
\subsection{Baseline Methods} 
The following sections provide details on the methodology, experimental setup. 

\textbf{Traditional ML Baselines.} We use Support Vector Machines (SVMs),  logistic regression, and Gaussian Naive Bayes models. SVM and logistic regression, commonly used for text classification with manually designed features, serve as strong benchmarks across various tasks. Despite their simplicity, they remain effective in handling noisy text from social media~\cite{ccoltekin2018tubingen, mohammad2018semeval}.  However, they face challenges with out-of-vocabulary terms, subtle semantic differences, and class imbalance. Gaussian Naive Bayes, a computationally efficient classification algorithm, performs well with high-dimensional data and minimal training data. However, it assumes feature independence, which may not hold in real-world scenarios. Additionally, model performance is sensitive to deviations in feature distribution from the assumed Gaussian distribution. In our baseline setting, we convert text data into numerical representation using Term Frequency-Inverse Document Frequency (TF-IDF) and use frequency-based features with classifiers from the Scikit-Learn library.

\textbf{Transformer Models.} We fine-tune BERT, DistilBERT, Mental BERT, Flan-T5, XLNet, and GPT-2.0 for multi-class classification. \textit{BERT}~\cite{devlin2019bert}, though successful in NLP, requires significant storage and computational power, making deployment challenging. Knowledge distillation compresses deep neural networks, creating smaller models with robustness. \textit{DistilBERT}~\cite{sanh2019distilbert} is a faster and more efficient alternative to BERT that maintains strong language comprehension. \textit{MentalBERT}~\cite{ji2022mentalbert}, a specialized pretrained language model, improves mental health applications by detecting mental disorders and suicidal ideation. It leverages contextualized language representations to enhance NLP models in mental healthcare tasks, outperforming general-purpose language models. \textit{Flan-T5}~\cite{chung2024scaling}, an instruction-tuned variant of T5, learns from diverse instruction-based datasets to perform well on various NLP tasks. The Flan-T5 base version, with 250 million parameters, strikes a balance between efficiency and capability, suitable for text generation and understanding tasks with a smaller computational footprint. \textit{XLNet}~\cite{yang2019xlnet}, is an enhanced version of BERT. Unlike BERT, which uses masked language modeling, XLNet employs Transformer-XL as its feature extraction architecture, introducing recurrence to capture long-range dependencies and develop a deeper understanding of language context. This makes it a powerful alternative for various NLP applications.  We use GPT-2.0~\cite{radford2019language}, an autoregressive model belonging to the GPT family to provide a broader comparative framework and evaluate its performance in conjunction with other pretrained models within our evaluation. We fine-tune multiple transformer-based classifiers with specific hyperparameters for text classification. BERT, DistilBERT, and MentalBERT use a learning rate of 1e-3, a batch size of 16, and 10 epochs. Flan-T5 is trained with a lower learning rate of 3e-4, a batch size of 8, and 10 epochs. XLNet follows the same learning rate of 1e-3 but with a batch size of 8. GPT-2.0 is fine-tuned with a learning rate of 3e-4, a batch size of 4, and 10 epochs.
\subsection{Experiment Results}
This section presents the performance of our models, compares baseline methods using key evaluation metrics, and highlights class-specific challenges. Table~\ref{tab:Comparison} illustrates the precision (P), recall (R), F-score (F), and accuracy (Acc) averaged over 10 folds, providing a comprehensive assessment of model effectiveness. MentalBERT consistently performed well across all classes, particularly in VA (0.87), SA (0.83), and PA (0.80). GPT-2.0 had strong performance in PA (0.78) and SA (0.78) but struggled with EA (0.36). DistilBERT showed balanced performance across categories but struggled in EA (0.40). Traditional ML models performed poorly on IA, VA, and EA, indicating difficulty in capturing semantic meaning.
EA has the lowest scores across all models, especially in traditional ML models. Even top-performing models like MentalBERT (0.48) and GPT-2.0 (0.36) struggle here, suggesting a need for data augmentation or alternative representations.
SpiA also sees lower F1 scores across models, particularly in Gaussian NB (0.30) and Linear SVM (0.32), suggesting challenges in capturing nuanced meanings.
\textit{For best overall performance:} MentalBERT is the top choice. \textit{For a balance of efficiency and performance:} DistilBERT and GPT-2.0 offer strong performance with potentially lower computational costs.
\begin{table*}[htbp]
\caption{Comparisons of baseline methods. Precision (P), and Recall (R) and F-score (F) are averaged over 10 folds.}
\scriptsize
\begin{center}
\begin{tabular}{|c|c c c|c c c|c c c|c c c|c c c|c c c|c|}
\toprule
\textbf{Method} & \multicolumn{3}{|c|}{\textbf{IA}} & \multicolumn{3}{|c|}{\textbf{VA}} & \multicolumn{3}{|c|}{\textbf{SpiA}} & \multicolumn{3}{|c|}{\textbf{PA}} & \multicolumn{3}{|c|}{\textbf{SA}} & \multicolumn{3}{|c|}{\textbf{EA}} &\textbf{Acc}\\
\cline{2-19} 
 & \textbf{P} & \textbf{R} & \textbf{F} & \textbf{P} & \textbf{R} & \textbf{F}  & \textbf{P} & \textbf{R} & \textbf{F} & \textbf{P} & \textbf{R} & \textbf{F}  & \textbf{P} & \textbf{R} & \textbf{F} & \textbf{P} & \textbf{R} & \textbf{F} &\\
\toprule
\textbf{LR} &0.71 &0.15 &0.25 &0.89 &0.53 &0.67 &0.31 &0.26 &0.29 &0.64 &0.75 &0.69 &0.50 &0.76 &0.60 &0.23 &0.17 &0.21 &\textbf{0.52}\\
\hline
\textbf{Linear SVM} &0.40 &0.24 &0.30 &0.73 &0.59 &0.66 &0.32 &0.32 &0.32 &0.67 &0.73 &0.70 &0.51 &0.65 &0.57 &0.20 &0.15 &0.17 &0.50\\
\hline
\textbf{Gaussian NB} &0.24 &0.24 &0.24 &0.21 &0.25 &0.23 &0.22 &0.50 &0.30 &0.64 &0.39 &0.48 &0.56 &0.39 &0.38 &0.18 &0.23 &0.20 &0.32\\
\toprule
\textbf{BERT} &0.41 &0.47 &0.44 &0.77 &0.87 &0.82 &0.38 &0.48 &0.43 &0.73 &0.74 &0.74 &0.83 &0.78 &0.81 &0.48 &0.33 &0.39 &0.65\\
\hline
\textbf{DistilBERT} &0.57 &0.63 &0.60 &0.70 &0.91 &0.79 &0.46 &0.67 &0.55 &0.79 &0.72 &0.76 &0.79 &0.84 &0.82 &0.75 &0.27 &0.40 &0.69\\
\hline
\textbf{MentalBERT} &0.70 &0.74 &0.72 &0.84 &0.91 &0.87 &0.63 &0.44 &0.52 &0.75 &0.85 &0.80 &0.77 &0.91 &0.83 &0.62 &0.39 &0.48 &\textbf{0.74}\\
\hline
\textbf{Flan-T5} &0.70 &0.37 &0.48 &0.69 &0.87 &0.77 &0.42 &0.48 &0.45 &0.75 &0.70 &0.73 &0.73 &0.84 &0.78 &0.44 &0.33 &0.38 &0.65\\
\hline
\textbf{XLNet} &0.52 &0.79 &0.62 &0.79 &0.83 &0.81 &0.48 &0.44 &0.46 &0.75 &0.70 &0.73 &0.82 &0.66 &0.73 &0.33 &0.39 &0.36 &0.63\\
\hline
\textbf{GPT-2.0} &0.60 &0.47 &0.53 &0.69 &0.78 &0.73 &0.41 &0.48 &0.44 &0.87 &0.70 &0.78 &0.67 &0.94 &0.78 &0.67 &0.24 &0.36 &0.66\\
\bottomrule

\end{tabular}
\label{tab:Comparison}
\end{center}
\end{table*}

\textbf{Explainability Evaluation.} We evaluate the top-performing models using LIME~\footnote{https://github.com/marcotcr/lime} for explainability analysis~\cite{saxena2022explainable} and calculate the similarity score between the LIME-generated predictions and the annotated explanation spans using keywords. To assess performance, we use ROUGE~\cite{lin2004rouge}, BLEU~\cite{papineni2002bleu}, F1-score, Precision, and Recall as evaluation metrics. As shown in Table~\ref{Tab:LIME}, analysis indicates that MentalBERT outperforms LR across all metrics, demonstrating superior accuracy and interpretability.
\begin{table}[htbp]
\caption{Explainability of Top performing models (LR and Fine-Tuned MentalBERT) using LIME}
\label{Tab:LIME}
\begin{center}
\begin{tabular}{|c|c|c|c|c|c|}
\hline 
\textbf{Method} &\textbf{F1-score} &\textbf{Precision} & \textbf{Recall} &\textbf{ROUGE} & \textbf{BLEU}\\
\hline
\textbf{LR} &0.4221 &0.314 &0.6976 &0.3645 &0.1349\\
\hline
\textbf{MentalBERT} &0.4471 &0.4901 &0.7463 &0.3833 &0.1412\\
\hline
\end{tabular}
\end{center}
\end{table}

%% file: LimConclusion.tex
\section{Limitations}
One of the key limitations of this work is the difficulty in accurately identifying emotional and spiritual dimensions due to their inherently subjective nature. This subjectivity often leads to varied interpretations of the same text by different annotators. For example, the phrase \textit{``I don't belong anywhere"} might be categorized as a Social Aspect by one annotator, while another might consider it an Emotional Aspect. Similarly, a statement like \textit{``I feel lost"} could be seen as reflecting a Spiritual Aspect by one annotator, indicating a search for meaning, whereas another might interpret it as an Emotional Aspect, suggesting feelings of confusion or distress. These discrepancies highlight the lack of a human factor in determining mental health, making it challenging to categorize wellness dimensions accurately. Addressing these overlaps and subjective interpretations is crucial for refining the classification process and ensuring more accurate and consistent results.
\section{Conclusion} We present an annotation framework for classifying wellness dimensions using mental health online forum data from the Australian region and introduce the Holistix dataset, developed with domain-specific guidelines and expert feedback. We compare traditional machine learning and transformer models for multi-class classification, using precision, recall, and F1 score as key evaluation metrics. The results highlight the performance of the models, along with challenges observed in specific wellness dimension classes. We evaluated model performance and applied LIME for post-hoc interpretability to enhance result transparency. Future work will explore multi-label classification to better handle overlapping wellness dimensions and conduct impact analysis to understand how different aspects interact. Additionally, we plan to incorporate large language models and leverage explanation span predictions to further enhance model explainability and improve overall classification accuracy.

%% file: EthicalStmt.tex
\section{Ethical Statement}
In conducting this study, we ensured strict adherence to ethical principles. All data were sourced from publicly accessible online forums that do not disclose user metadata, thereby safeguarding individual privacy. To further protect user identities, we anonymized any potentially identifying information within the dataset. Recognizing the inherent subjectivity in data annotation, we implemented comprehensive guidelines to minimize bias and enhance the reliability of our labeled data. These guidelines were developed in collaboration with domain experts and included clear instructions for annotators. High inter-annotator agreement scores indicate the effectiveness of these measures. To promote transparency and reproducibility, our dataset and source code are publicly available on GitHub, enabling other researchers to replicate our baseline results. Through these efforts, we affirm our commitment to upholding ethical standards in research.